\documentclass[journal,twoside,web]{IEEEtran}
\pdfoutput=1
\usepackage{cite}
\usepackage{amsmath,amssymb,amsfonts}
\usepackage{siunitx}
\usepackage{algorithmic}
\usepackage{graphicx}
\usepackage{textcomp}

\begin{document}

\title{Predicting dynamic, motion-related changes in $\text{B}_{\text{0}}$ field in the brain at a $\text{7}\,\text{T}$ MRI using a subject-specific fine-tuned U-net}

\author{Stanislav Motyka, Paul Weiser, Beata Bachrata, Lukas Hingerl, Bernhard Strasser, Gilbert Hangel, Eva Niess, Dario Goranovic, Fabian Niess, Maxim Zaitsev, Simon Daniel Robinson, Georg Langs, Siegfried Trattnig, Wolfgang Bogner

\thanks{S.M., P.W., B.B., L.H., B.S., G.H., E.N., D.G., F.N., S.R., S.T, and W.B. are with High Field MR Center, Department of Biomedical Imaging and Image-guided Therapy, Medical University of Vienna, Vienna, Austria, email: wolfgang.bogner@meduniwien.ac.at}

\thanks{S.T. and W.B. are with Christian Doppler Laboratory for Clinical Molecular MR Imaging, Medical University of Vienna, Vienna, Austria; G.H. is with Department of Neurosurgery, Medical University of Vienna, Vienna, Austria; M.Z. is with Department of Radiology – Medical Physics, University of Freiburg, Faculty of Medicine, University of Freiburg – Medical Centre, Freiburg, Germany; P.W. and G.L. are with Computational Imaging Research Lab, Department of Biomedical Imaging and Image-guided Therapy, Medical University of Vienna, Vienna, Austria; B.B. and S.T. are with Karl Landsteiner Institute for Clinical Molecular MR in Musculoskeletal Imaging, Vienna, Austria; B.B. is with Department of Medical Engineering, Carinthia University of Applied Sciences, Klagenfurt, Austria}
\thanks{The financial support by the Austrian Science Fund (FWF P 34198 and FWF TAI-676) is gratefully acknowledged.}}

\maketitle

\begin{abstract}
Subject movement during the magnetic resonance examination is inevitable and causes not only image artefacts but also deteriorates the homogeneity of the main magnetic field ($B_0$), which is a prerequisite for high quality data. Thus, characterization of changes to $B_0$, e.g. induced by patient movement, is important for MR applications that are prone to $B_0$ inhomogeneities. We propose a deep learning based method to predict such changes within the brain from the change of the head position to facilitate retrospective or even real-time correction. A 3D U-net was trained on \textit{in vivo} brain $\SI{7}{T}$ MRI data. The input consisted of $B_0$ maps and anatomical images at an initial position, and anatomical images at a different head position (obtained by applying a rigid-body transformation on the initial anatomical image). The output consisted of $B_0$ maps at the new head positions. We further fine-tuned the network weights to each subject by measuring a limited number of head positions of the given subject, and trained the U-net with these data. Our approach was compared to established dynamic $B_0$ field mapping via interleaved navigators, which suffer from limited spatial resolution and the need for undesirable sequence modifications. Qualitative and quantitative comparison showed similar performance between an interleaved navigator-equivalent method and proposed method. We therefore conclude that it is feasible to predict $B_0$ maps from rigid subject movement and, when combined with external tracking hardware, this information could be used to improve the quality of magnetic resonance acquisitions without the use of navigators.

\end{abstract}


\section{Introduction}
\label{sec:introduction}
\IEEEPARstart{A}{ll} \textit{in vivo} Magnetic Resonance Imaging (MRI) examinations are sensitive to subject motion. Those requiring prolonged measurement times, are particularly susceptible \cite{Heckova2018, Zaitsev2015}. A change in the subject’s position causes motion artefacts and decreases the homogeneity of the static magnetic field ($B_0$) \cite{Bogner2014, Liu2018}. Changes in $B_0$ are increasingly pronounced at ultra-high-field MR scanners ($B_0 \geq \SI{7}{\tesla})$ \cite{Juchem2017}.

A spatially homogeneous—or at least temporarily stable—$B_0$ field is a prerequisite for several MRI methods. For instance, in MR spectroscopy (MRS), intra-voxel $B_0$ inhomogeneities and temporal frequency changes degrade the spectral resolution, which translates into reduced chemical specificity \cite{Motyka2019}. In MRS imaging, they aggravate artefacts arising from extracranial lipid and unsuppressed water signals \cite{Maudsley2020}. In fast imaging, $B_0$ inhomogeneities cause nonlinear image distortions (e.g., for echo planar imaging) or image blurring (e.g., for spiral acquisitions) \cite{Wu2008}. For Chemical exchange saturation transfer (CEST), $B_0$ inhomogeneities induce frequency offsets \cite{Sun2007} which cause systematic errors in quantification. Long MRI sequences or those with many repetitions are even more vulnerable to subject motion. To tackle those issues, several MR-based and external tracking methods have been proposed, which provide information about the change of the patient position, and some of them are capable to map the $B_0$ distribution.

MR-based tracking methods consist of short MR sequences (termed navigators) for example based on fast gradient echo scans with echo-planar imaging (EPI) read-out \cite{Andronesi2021}. Navigators are temporally interleaved with the main (parent) sequence that requires correction \cite{White2010}. Simple navigators can monitor subject position and more advanced volumetric navigators (vNavs) can even map changes of the $B_0$ field over time \cite{Hess2011, Moser2020}. However, for vNavs, the acquisition alone can be as long as $\SI{500}{ms}$ \cite{Liu2020} and can thus not be easily inserted into the majority of sequences, especially not those with short TR (frequently under $\SI{10}{ms}$) \cite{Andronesi2021}. Simpler navigators are easier to implement, but they can only measure global frequency drift and cannot capture the spatial distribution of $B_0$ changes \cite{Bogner2014}. Finally, self-navigation allows motion to be monitored e.g. based on repeated resampling of the k-space center via the parent sequence. Self-navigation does not require additional scans, but reduces the SNR efficiency of the sequence and has limited or no ability to characterize $B_0$ field changes depending on the contrast of the main sequence \cite{Kim2004}. 

External tracking methods use additional hardware. Their advantage over navigators is that motion (detected for example by optical tracking \cite{Zaitsev2006,Maclaren2012}) or changes of the $B_0$ field (detected for example by NMR probes \cite{Zanche2008}) are acquired independently from the MR scanner and are thus compatible with every sequence. However, optical or similar tracking systems do not provide information about $B_0$ field changes. NMR probes can track the $B_0$ field, but only outside the subject, and translating this to accurate $B_0$ estimates inside the body is challenging or needs to be combined with conventional $B_0$ mapping \cite{Barmet2008}. While optical tracking is already established as a clinically approved commercial product, an NMR probe system is a highly specialized and costly piece of equipment that is not generally supplied as part of an MRI system.

Thus, an approach that combines the benefits of external motion tracking (i.e., independence from the parent sequence, highly accurate tracking with high temporal resolution) with that of internal navigators (i.e., accurate dynamic volumetric $B_0$ mapping) without their disadvantages, is highly desirable. Ideally, it would allow improved real-time (or retrospective) correction of both motion and $B_0$ instabilities.

In recent years, deep learning methods have proved to be successful in uncovering hidden patterns in image data, which can be leveraged to solve complex problems, provided that sufficient training data are available \cite{LeCun2015}. For MRI methods, the image reconstruction \cite{Knoll2020} as well as the segmentation \cite{Liu2021} and many other problems \cite{Lundervold2019} can be potentially overcome by deep learning-based methods. 

In this study, we propose a neural network (NN) approach to predict changes of the $B_0$ field within the brain from observed changes of the head position and orientation. A U-net is used to predict a $B_0$ map from the following input: (i) anatomical MRI at the initial position, (ii) initial $B_0$ map, and (iii) head pose change at a certain time point described via six degrees of freedom. The $B_0$ maps are predicted for each known head position/time point. A general set of weights of the U-net is estimated using the data of 11 volunteers. These weights are then fine-tuned for each volunteer using the acquired $B_0$-maps of six head positions for that volunteer to include subject-specific information and thus improve the $B_0$-prediction. The whole proposed method would therefore include the measurement of an anatomical MRI sequence, and the $B_0$-sequence of six head positions to refine the general network weights for the specific subject in a short training ($\sim \SI{1}{\text{min}}1$) while the subject is in the scanner. These weights are then used to predict the $B_0$-changes caused by motion, which can be used to correct the data.
 
\begin{figure}
	\centering
		\includegraphics[trim=0 0 0 0,clip,width=\linewidth]{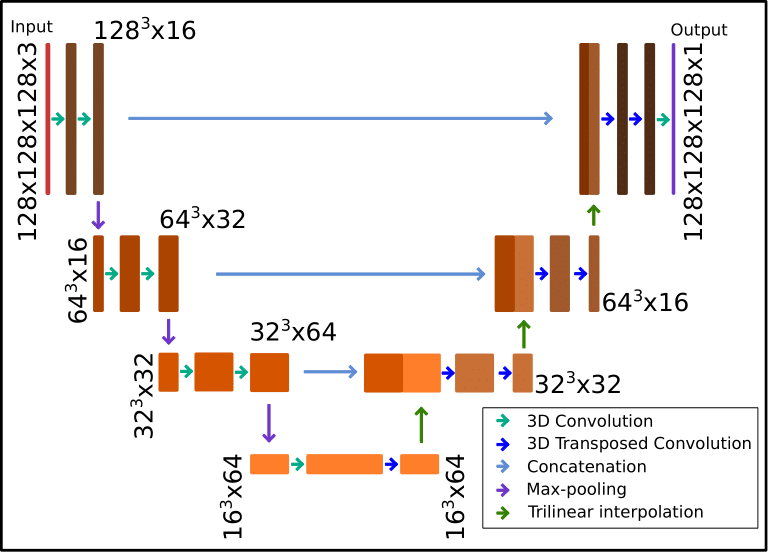}
	\caption{3D U-net architecture used in the study. The input to the network has three features: (i) $B_0$ map of the initial position, (ii) Anatomical reference of the initial position, and (iii) Anatomical reference of a new position. The output has one feature: $B_0$ map of the new position. }
	\label{fig:3DunetArch}
\end{figure}

\section{Methods}
\label{sec:methods}

\subsection{Experimental data}
All measurements were carried out on a $\SI{7}{T}$ Magnetom+ MR Scanner (Siemens Healthineers, Erlangen, Germany) with a 32-channel head coil (Nova Medical, Wilmington, MA). 15 healthy volunteers (11 males and 4 females) were included in this study. The study was approved by the Ethics Committee of the Medical University of Vienna and written informed consent was obtained from all volunteers.

For each volunteer, an MP2RAGE image \cite{Marques2013} was acquired as anatomical reference with nominal resolution of $1.1 \times 1.1\times1.\SI{1}{mm}$, FOV $=220\times220\times\SI{220}{mm}$, TE/TR $= 3.28/\SI{5000}{ms}$, TI $= \SI{700}{ms}$, TI2 $=\SI{2700}{ms}$, GRAPPA factor $= 4$, TA = $4:\SI{57}{mins}$.

The $B_0$ maps were acquired at 30 random head positions per volunteer. All volunteers were asked to change their head positions randomly, to cover the possible range within the head coil. The first head position was identical to that for the MP2RAGE scans. At each head position, two sequences for $B_0$ mapping were run: 

(i) 2D multi-echo gradient echo (GRE) sequence with nominal resolution of $1.9\times1.\SI{9}{mm}$, FOV $=240\times\SI{240}{mm}$, GRAPPA factor $= 4$, $80$ slices with $\SI{2}{mm}$ thickness, TR $= \SI{1410}{ms}$, $\text{TE}_{1-5} = 3/6/9/12/\SI{15}{ms}$, Flip angle $= 55^{\circ}$, TA $=\SI{59}{s}$; 

(ii) 3D dual-echo echo planar imaging (EPI) sequence with nominal resolution of $8.0\times8.0\times\SI{8.0}{mm}$, FOV $=256\times\SI{256}{mm}$, $32$ slices, EPI factor $= 16$, TR $= \SI{9.0}{ms}$, $\text{TE}_{1-2}  = 3.8/\SI{4.8}{ms}$, Flip angle $= 2^{\circ}$, TA $= \SI{0.6}{s}$. 

\begin{figure*}
	\centering
	\includegraphics[trim=0 0 0 0,clip,width=\linewidth]{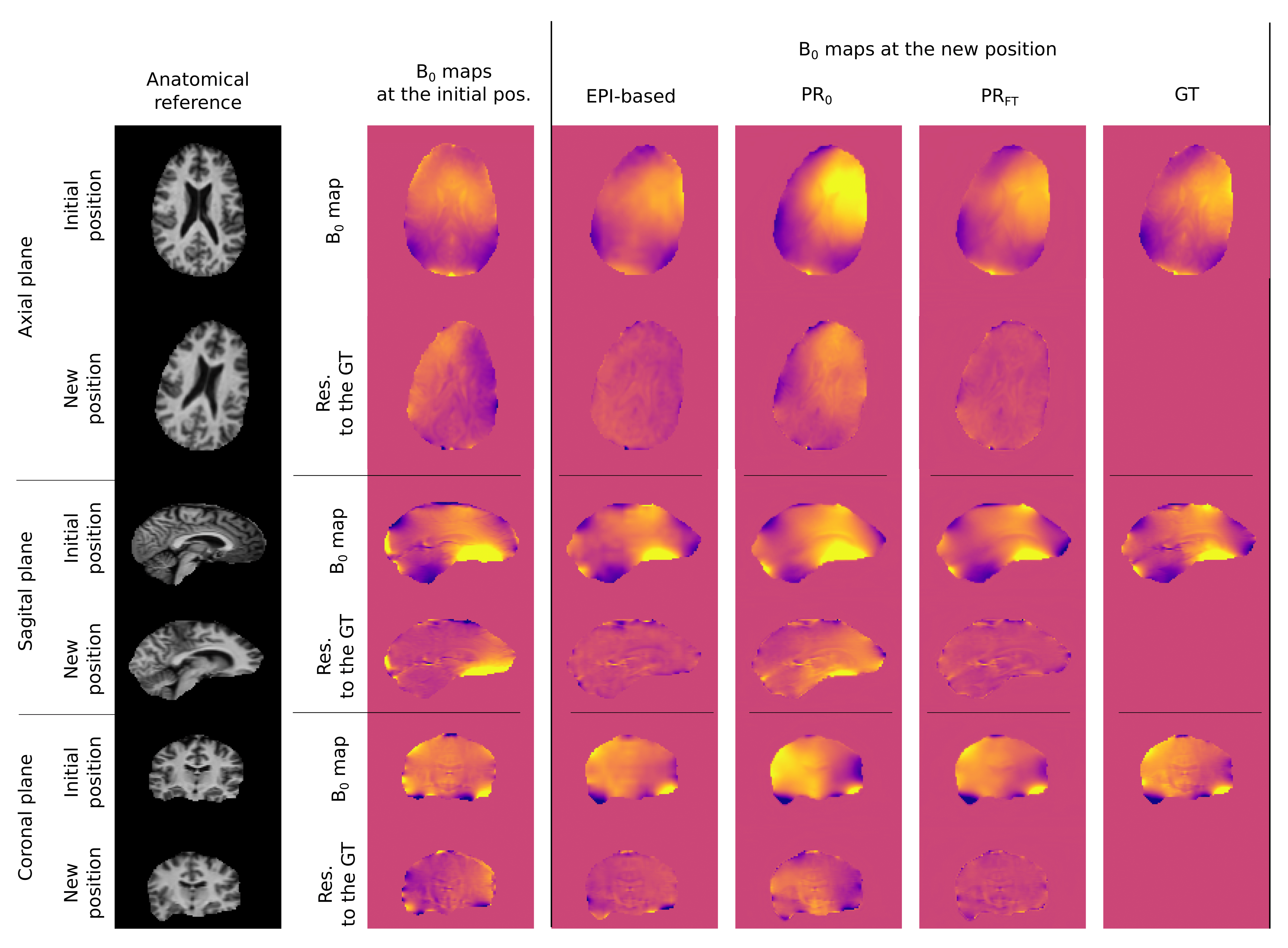}
	\caption{Comparison of $B_0$ maps of four approaches. On the very left, the anatomical reference is depicted of one volunteer at the initial and the new head position. For both positions three slices from three planes are plotted. The second column consists of $B_0$ maps at the initial position and the residua to the ground truth $B_0$ map of the new position. The next four columns depict $B_0$ maps of three approaches: (i) EPI-based approach (EPI), (ii) predicted $B_0$ map from not fine-tuned NN (PR\textsubscript{0}), (iii) predicted $B_0$ map from fine-tuned NN (PR\textsubscript{FT}); and the ground truth $B_0$ maps (GT) plus their residua to the ground truth $B_0$ map at the new position.}
	\label{fig:b0maps}
\end{figure*}

\subsection{Experimental data for physics-driven augmentation}

In a separate experiment, measurements with a spherical phantom were used to map the 1st and the 2nd order spherical harmonics of the shimming system of the $\SI{7}{T}$ Magnetom+ MR Scanner via the same multi-echo GRE sequence described above. $B_0$ shimming was performed using the standard automatic shim procedure and the initial $B_0$ map was measured. The current amplitudes for each spherical harmonic term were manually altered four times from its initial $B_0$ shim setting in a linear fashion ($\SI{-100}{\mu T/m^n}$, $\SI{-50}{\mu T/m^n}$, $\SI{+50}{\mu T/m^n}$ and $\SI{+100}{\mu T/m^n}$, where $n$ is the order of the spherical harmonic). After each modification, another $B_0$ map was acquired. These data were later used for data augmentation in the neural network training.

\subsection{Pre-processing of experimental data}

For each head position, $B_0$ maps were calculated from the GRE sequence and the EPI-based sequence. GRE-based $B_0$ maps were calculated from the magnitude and phase images coil combined by ASPIRE \cite{Eckstein2018}, and phase unwrapped using ROMEO \cite{Dymerska2021}. $B_0$ maps from the 2TE-EPI sequence were calculated as Hermitian inner product \cite{Bernstein1994}. GRE-based $B_0$ maps were acquired in high spatial resolution with multiple echo time, thus served as the gold standard method to estimate $B_0$ inhomogeneity. For the neural network training, GRE-based $B_0$ maps were considered the ground truth. Low spatial resolution EPI-based $B_0$ maps are equivalent to the dual-echo navigators, which can be used to estimated $B_0$ inhomogeneity in the dead time of the parent sequence \cite{Hess2011}.

MP2RAGE data were transformed to different head positions using Flirt from the FSL toolbox \cite{Jenkinson2002} by applying transformation matrices from the co-registration of the first position to the other positions using the magnitudes of the first echoes. MP2RAGE datasets were also used to calculate brain masks (BET, FSL toolbox \cite{Jenkinson2002}), which were transformed in the same way as described above. 

Spherical harmonics of the the first and the second order of the shimming system (i.e.X, Y, Z, XY, ZY, Z2, ZX, X2-Y2) \cite{Juchem2017} were characterized using the five $B_0$ maps with different shim current amplitudes. The measured $B_0$ field associated with each shim term was fitted with the respective analytical spherical harmonic function \cite{Caola1978} by a nonlinear curve fitting solver in Matlab \cite{matlab17}.

The training dataset consisted of the 319 instances from 11 volunteers. Each instance contained the input: (i) anatomical MRI (i.e., T1-weighted MP2RAGE) at the initial position, (ii) $B_0$ map at initial position, and (iii) the same anatomical MRI, but after applying the 6DoF transformation to the new position. The output consisted of the $B_0$ map at the new position. 

Augmentation of the training dataset was performed during each epoch of the network training by adding the same, randomly-scaled spherical harmonic $B_0$ fields to both input and output $B_0$ maps of one instance. The spherical harmonics are normally used for $B_0$ shimming of the volume-of-interest. Thus, that data augmentation is physically meaningful since the changes of the $B_0$-maps with motion should not depend on the $B_0$ shimming, and by adding spherical harmonics $B_0$ fields, the acquisition of the same volunteer under different shimming conditions is simulated. 

The test dataset consisted of the data of 4 volunteers. The pre-processing was similar to that performed on the training dataset.

\subsection{Architecture and training of neural network}

All calculations were performed on a DGX station equipped with Tesla V100 GPU cards (Nvidia, Santa Clara, CA, US). The PyTorch DL framework \cite{Paszke2017} was used. 

The U-net architecture was used \cite{Ronneberger2015} because of the ability to extract features from the input data at different spatial resolutions which are later used in the decoder part of the network to form a prediction. The network had 4 levels, with the encoder part at each level containing the two 3D convolution layers each followed by the leakyRelu activation. The spatial resolution was decreased with a max-pooling layer by a factor of two. The bottom of the U-net consists of two 3D convolution and one 3D transposed convolution layers, each followed by leakyRelu activation. The decoder part, at each level, consists of two 3D transposed convolution each followed by leaky Relu. The spatial resolution was increase by a trilinear interpolation by factor of 2. All convolutional layers had a kernel size of 5 in all three spatial dimensions and convolutions were performed with a stride of one. The skipped connection was performed as a concatenation of features from the encoder part to the features of the same spatial resolution in the decoder part. At the end, 3D convolution was performed with a kernel size of one. The architecture is depicted in Figure \ref{fig:3DunetArch}.

The training was performed for 2000 epochs with a mini batch of 10. The Adam optimizer was used with a learning rate of 1e-5 and weight decay of 1e-7. For each epoch during the training, the order of the training dataset was randomly permuted and each instance was augmented by the randomly scaled spherical harmonics. The mean-squared error of the unwrapped $B_0$ map and the prediction formed a loss function.

\subsection{Fine-tuning to specific subject}

The U-net trained on the training dataset was fine-tuned to each subject with a very short training (50 epochs). The first six head positions from all volunteers in the test dataset were separated for the fine-tuning training and the 23 head position were kept for evaluation. The Adam optimizer was used with learning rate of 1e-6, and weight decay of 1e-7.

\subsection{Evaluation}

The accuracy of the subject-specific, fined-tuned U-net (PR\textsubscript{FT}) was compared to three approaches: (i) no-correction (NC), for which the $B_0$ map was not updated and directly compared to the initial head position; (ii) prediction of NN, which was not fine-tuned (PR\textsubscript{0}); (iii) EPI-based approach (EPI), in which the $B_0$ maps were measured at the new position with a navigator-like sequence set up. The $B_0$ maps of all four approaches were compared against the ground truth data (GRE-based $B_0$ maps) and residua maps were calculated. 

The $B_0$ maps and the residua maps were compared qualitatively. The quantitative analysis was performed using the absolute values of residua maps within the brain mask. For each head position in the testing dataset, median and interquartile range was calculated of the difference to the ground truth. Boxplots of these values were created, which summarize the approach overall performance on the test dataset. 

The fine-tuning of the NN for a specific subject was analyzed in terms of the required minimum number of head positions used for fine-tuning training and the number of epochs. Fine-tuning was tested with 3, 4, 5, and 6 head positions and in each case the fine-tuning was performed for 50 epochs. The number of epochs was tested with 6 head position and 5, 10, 20, 35, 50, 75, 100, 150 and 200 epochs were tested. The quantitative analysis was run over the residua maps of each fine-tuning test in the same fashion as describe above.

\subsection{Reproducible research}

The code is available on Github [\textit{link to the repository will be avaliable}].

\section{Results}

\subsection{Accuracy of network prediction}

The qualitative comparison of $B_0$ maps of 4 approaches and their residua to the ground truth $B_0$ maps are depicted along with the anatomical references in Figure \ref{fig:b0maps}. The results are presented in three orthogonal planes. In the axial and coronal planes, residua maps between the initial and the new position show a clear left-right gradient of the error, which is caused by patient movement. In the sagittal plane, a high amplitude hotspot of error is visible in the frontal lobe.

The gradient as well as the hotspot in the frontal lobe are not visible for the EPI approach and PR\textsubscript{FT}. The residua maps of the PR\textsubscript{0} contain the slow spatial gradient in all three orthogonal planes.    

\begin{figure}
	\centering
	\includegraphics[trim=0 0 0 0,clip,width=\linewidth]{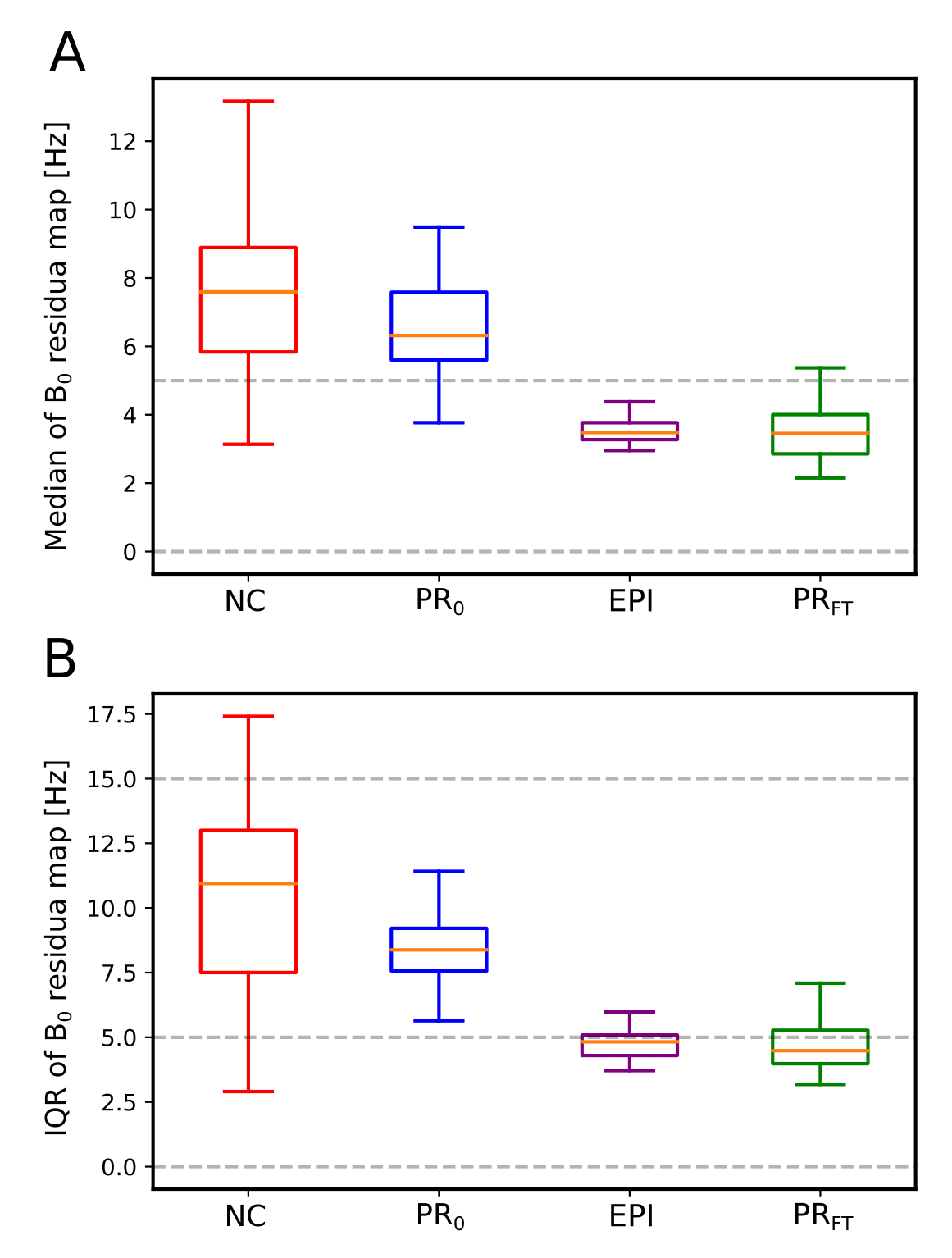}
	\caption{Comparison of the medians (section A) and the IQRs (section B) of the residua maps between four tested apporaches and the ground truth. The apporachers are: No-correction (NC), prediction by the non-fine-tuned NN (PR\textsubscript{0}), EPI-based $B_0$ mapping (EPI), and prediction by the fine-tuned NN (PR\textsubscript{FT}).}
	\label{fig:b0maps_bxp_of_mean_and_IQR_b0residuum}
\end{figure}

\begin{figure*}
	\centering
        \includegraphics[trim=0 0 0 0,clip,width=\linewidth]{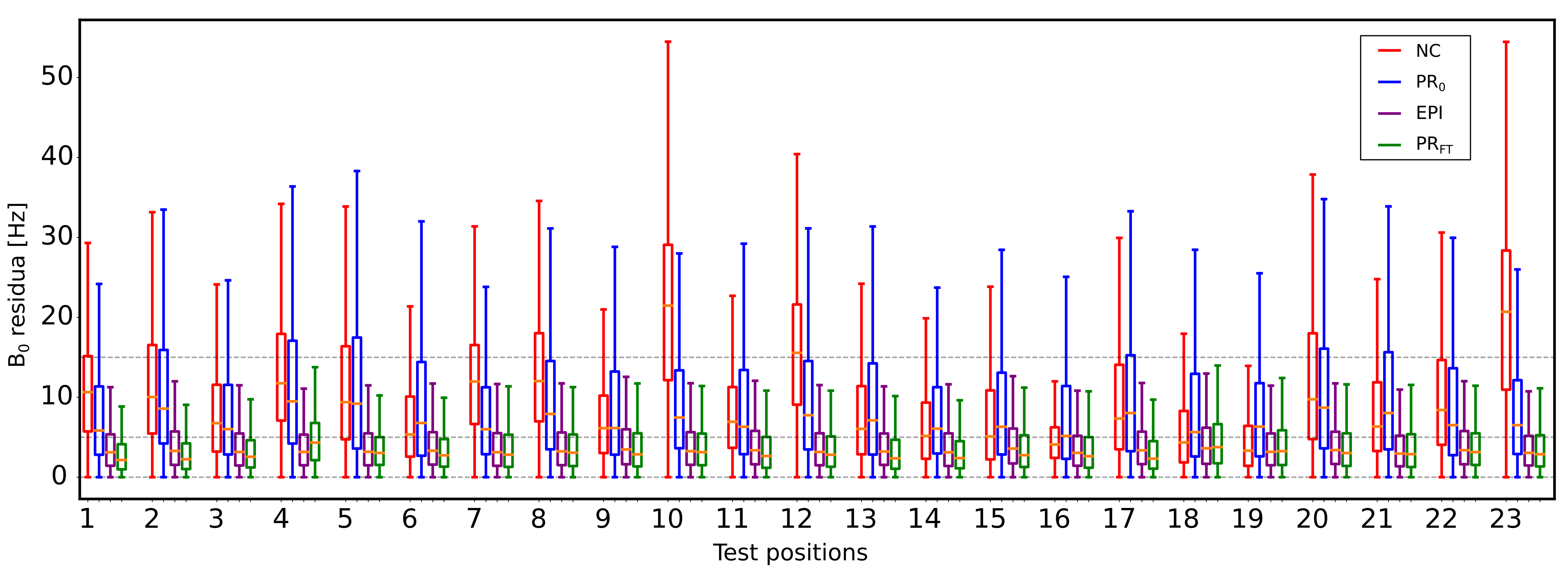}
        \caption{Quantitative results of one volunteer. The boxplots of the absolute $B_0$ residua to the ground truth for 4 tested approaches at 23 head positions are presented. No-correction (NC), Prediction by non-fine-tuned NN (PR\textsubscript{0}), EPI-based $B_0$ mapping (EPI) and prediction by fine-tuned NN (PR\textsubscript{FT}) are compared. The fine-tuning was performed with 6 brain volumes in 50 epochs.}
	\label{fig:b0maps_residua_per_position}
\end{figure*}

Quantitative comparison of overall performance of the four approaches is depicted in Figure \ref{fig:b0maps_bxp_of_mean_and_IQR_b0residuum}. The medians and the IQRs of the absolute values of residua maps are compared. The EPI approach as well as PR\textsubscript{FT} yield lower absolute residua compared to the NC approach. For the NC approach, the median of the median of the absolute residua was $\SI{7.59}{Hz}$, while for the EPI approach it was significantly lower, $\SI{3.48}{Hz}$ (p-value $<< 0.0001$), as well as for PR\textsubscript{FT}, $\SI{3.45}{Hz}$ (p-value $<< 0.0001$). There was no significance difference between the EPI approach and the PR\textsubscript{FT} (p-value $= 0.69$).

For the NC approach the median of the IQR of the absolute residua was $\SI{10.94}{Hz}$, which was significantly higher compared to the other three approaches: $\SI{8.37}{Hz}$ (p-value $= 2.21e-5$) for PR\textsubscript{0}; $\SI{4.82}{Hz}$ (p-value $<< 0.0001$) for EPI approach, $\SI{4.48}{Hz}$ (p-value $<< 0.0001$) for PR\textsubscript{FT}. The PR\textsubscript{FT} results are significantly lower to the PR\textsubscript{0} (p-value $<< 0.0001$). There was no significant difference between PR\textsubscript{FT} and the EPI approach (p-value $= 0.57$).

A quantitative comparison of methods for one volunteer is depicted in Figure \ref{fig:b0maps_residua_per_position}. 23 head positions are evaluated. Boxplots of absolute residua maps for each method at each head position are plotted. For the NC approach and the PR\textsubscript{0} the medians of absolute residua are above $\SI{5}{Hz}$ in all cases. For the EPI approach and the PR\textsubscript{FT}, the medians in all case are below 5 Hz. Moreover, the average of upper quartiles is 5.58 Hz for the EPI approach and 5.19 Hz for the PR\textsubscript{FT}.

\subsection{Analysis of fine-tuning procedure}

\begin{figure}
	\centering
	\includegraphics[trim=0 0 0 0,clip,width=\linewidth]{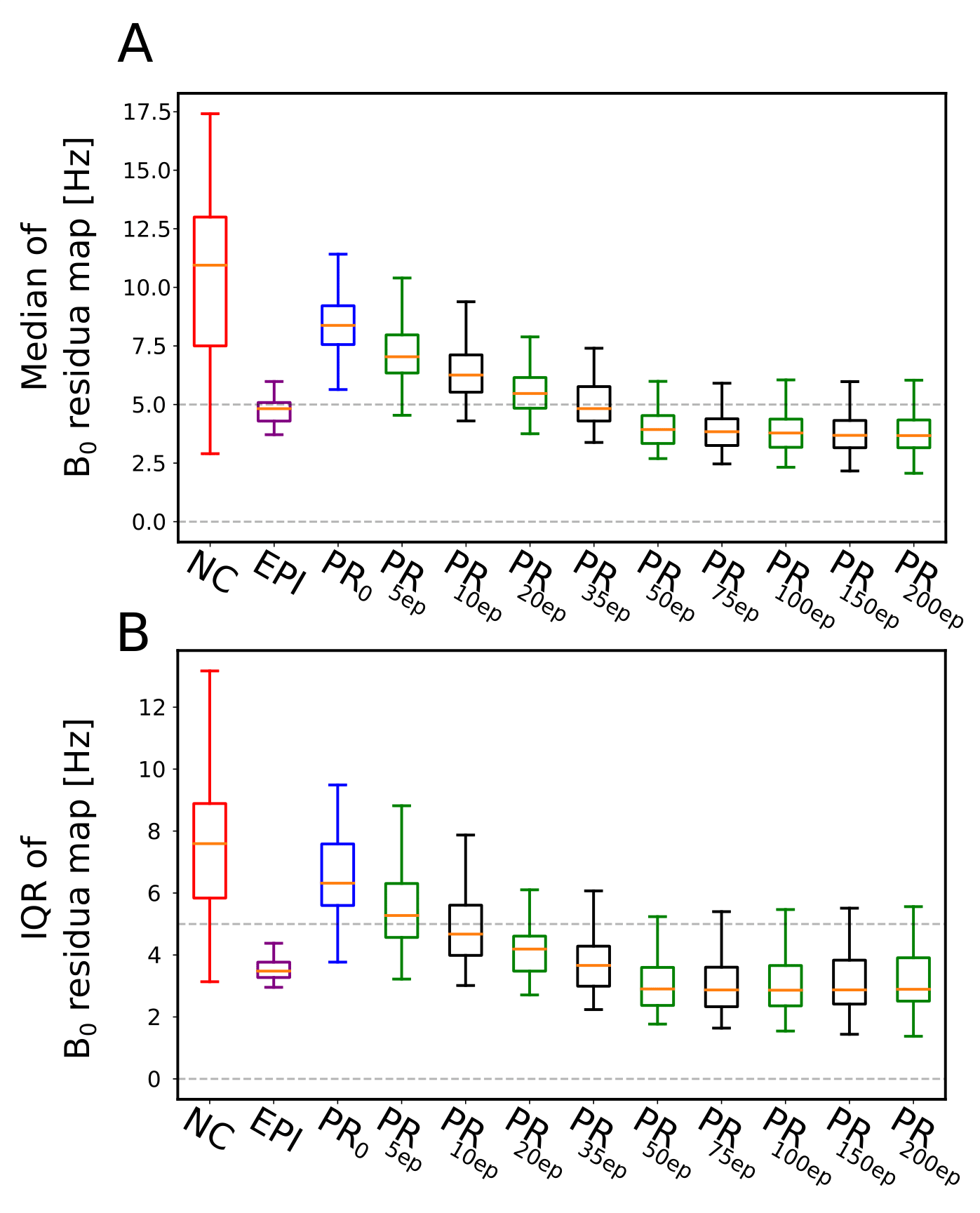}
	\caption{Quantitative comparison of the number of epochs for fine tuning training. A range of epoch was between 5 and 200. The results of fine-tuning are plotted along side other three methods: No-correction (NC), EPI-base $B_0$ mapping (EPI), and prediction by non-fine-tuned NN (PR\textsubscript{0}). Section A: Comparison of the medians of absolute residua maps. Section B: Comparison of the IQRs of the absolute residua maps.}
	\label{fig:FineTuning_number_of_epochs}
\end{figure}

\begin{figure}
	\centering
	\includegraphics[trim=0 0 0 0,clip,width=\linewidth]{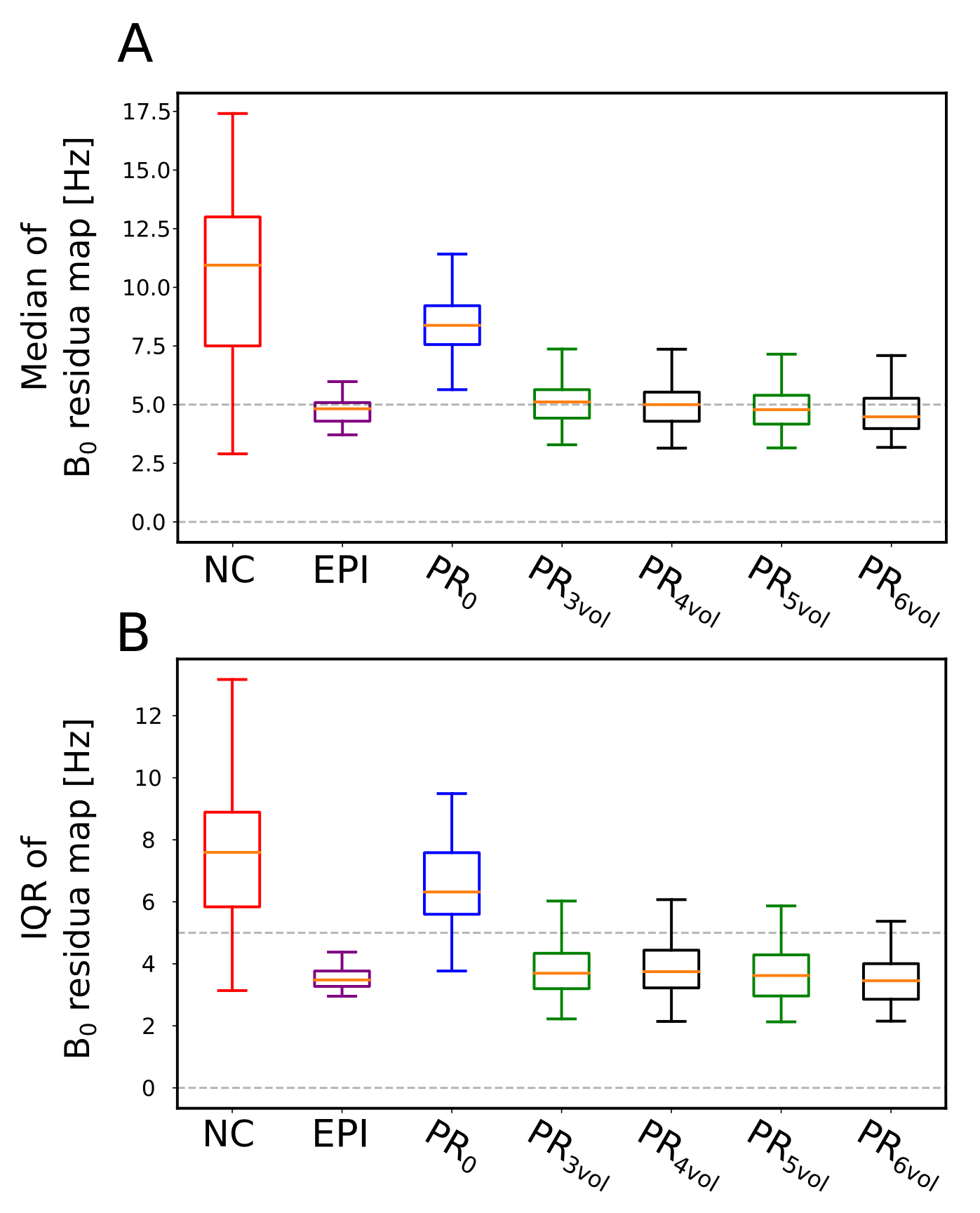}
	\caption{Quantitative comparison of the number of volumes for fine tuning training. The range of used volumes was between 3 and 6. The results of fine-tuning are plotted along side other three methods: No-correction (NC), Epi-base $B_0$ mapping (EPI), and prediction by non-fine-tuned NN (PR\textsubscript{0}). Section A: Comparison of the medians of absolute residua maps. Section B: Comparison of the IQRs of the absolute residua maps.}
	\label{fig:FineTuning_number_of_volumes}
\end{figure}

Quantitative results of the fine-tuning evaluation are depicted in Figure \ref{fig:FineTuning_number_of_epochs} in terms of number of epochs and in Figure \ref{fig:FineTuning_number_of_volumes} for the number of brain volumes used for the fine-tuning. 

The effect of using a different number of epochs for the fine-tuning was evaluated in a range between 5 and 200 epochs. In Figure \ref{fig:FineTuning_number_of_epochs}, section A, a trend of decreasing of the median of the absolute median residua can be observed in a range between 5 and 50 epochs. In a range between 50 and 200 epochs, the difference in the median values were not observed. 

The IQR metric followed the same trend, as shown in Figure \ref{fig:FineTuning_number_of_epochs}, section B. Only in the the range between 5 and 50 epoch of fine-tuning training, the values were decreasing. For the fine-tuning, which lasted longer then 50 epochs, there were no differences  compared to the case of 50 epochs of the fine-tuning. 

The number of volumes used for the fine-tuning training were compared in a range 3 to 6 brain volumes. There were no major differences for the median of the absolute median residua, depicted in Figure \ref{fig:FineTuning_number_of_volumes}, section A, nor for the IQRs of the absolute median residua, depicted in Figure \ref{fig:FineTuning_number_of_volumes}, section B. 

\section{Discussion}

This work presents the proof-of-principle investigation of a method to predict motion-induced $B_0$ changes via a NN from available rigid body head motion logs (e.g., obtained from external motion tracking), an initial $B_0$ map and an initial anatomical image. The prediction of the $B_0$ maps was carried out with a U-net trained with the experimentally acquired and augmented data of 11 volunteers. For each volunteer in the test dataset, the network was fine-tuned with a small number of subject-specific data and a limited number of epochs. The performance of the network was compared with three other approaches using a test dataset of four volunteers.

$B_0$ (as well as $B_1$) inhomogeneities can cause severe artifacts in reconstructed images. Several other methods were therefore proposed in the past to estimate field inhomogeneities and correct for them \cite{Robinson2011, Jezzard1995, Chen1999, Reber1998, Sekihara1985}. Such calibration methods typically require a specific calibration scan at the beginning of each MRI acquisition protocol. The results are then used to set the parameters of the following sequences. In case of subject movement, the $B_0$ information is not updated, which is analogous to the no-correction approach. Repeated $B_0$ mapping throughout the acquisition protocol is therefore necessary to account for any temporal instability (e.g., patient movement related $B_0$ changes). vNavs—typically interleaved with the main (parent) sequence—are able to provide dynamic $B_0$ estimates, but would often increase the total scan time and lower the SNR-per-unit-time efficiency in an unacceptable way \cite{Moser2020}. In some specific cases (e.g., fMRI), the $B_0$ maps can be calculated from the phase of EPI images \cite{Dymerska2018}, but such approaches are not very flexible and often require making certain compromises in the parent sequence \cite{Dymerska2016}. In contrast, our NN approach is capable to predict the change of the $B_0$ with high temporal resolution without measuring extra data during the subsequent sequences in the particular volunteer's MRI acquisitions. If motion information can be acquired externally, e.g. by optical tracking or directly from the MR data, no sequence parameter changes are necessary for our proposed NN method, which relies only on an accurate knowledge of the transformation matrices describing the rigid body motion. The fine-tuning procedure depends on the limited number of subject-specific data, which can be acquired at the beginning of the MR investigation protocol.

In recent years, deep learning methods have been applied in the reconstruction of MRI data \cite{Lundervold2019}, mainly assuming that the acquired k-space data are artefact-free. A small number of methods have been proposed which reduce or remove some artifacts arising from $B_0$ inhomogeneities but this is, to the best of our knowledge, the first which directly predicts $B_0$ map inhomogeneities. In the context of distortions in EPI images caused by $B_0$ inhomogeneities, deep-learning-based methods have been proposed to directly predict distortion free images \cite{Hu2020, Schilling2020}. Another deep-learning based approach was designed to compensate for the artifact due to $B_0$ fluctuations arising from respiration in multi-slice GRE by predicting the phase error term from the corrupted images \cite{An2021}. In contrast, our approach is less direct, but more flexible.

We have used subject-specific fined-tuned NN to predict $B_0$ maps. In general, this PR\textsubscript{FT} approach outperformed PR\textsubscript{0} (i.e., without fine-tuning) and the NC approach. The results of PR\textsubscript{FT} are similar to those obtained with the EPI approach. The NC method yielded the highest medians of $B_0$ residua as well as IQRs. The $B_0$ residua maps shows the left-right gradient of error in the axial and the coronal plane. The sagittal plane frequently shows error hotspots in the frontal lobe. Those effects are in agreement with a previously published analysis \cite{Hess2012}. 

The PR\textsubscript{0} approach yielded slightly better results compared to the NC approach. The quantitative comparison showed improvement in the medians and the IRQs of $B_0$ residua maps. From the investigation of methods per single head position, the PR\textsubscript{0} had similar performance for each position, no matter how severe the error of the NC approach was. The residua maps of the PR\textsubscript{0} shows a reverse gradient of the error compared to the NC approach. However, the $B_0$ maps themselves have similar features compared to the PR\textsubscript{FT}. The main difference is in their magnitude.

PR\textsubscript{FT} results were similar to those with the EPI approach in several investigations. The $B_0$ maps of both methods are comparable to the ground truth. Their $B_0$ maps residua did not contain the left-right gradient and the frontal lobe hotpot, which are typical $B_0$ inhomogeneities originating from subject movement. Quantitative results showed the same results for the median and IQRs of the absolute $B_0$ map residua in the overall comparison of the methods as well as in the comparison of methods per head volume.

The EPI approach and the fine-tuning required acquisition of additional subject-specific data. However, while the EPI approach acquires data during the whole scan, the data for fine-tuning are only acquired at the beginning of the MR protocol as a prescan. Once the network is fine-tuned for a specific subject, only the tracking of movement is required. These motion logs could be acquired via several internal or external methods \cite{Andronesi2021}. The one most independent from the MR acquisitions, and hence the most versatile, is optical tracking, in which information about the movement is sampled at a very high temporal rate up to $\SI{80}{Hz}$ \cite{Maclaren2012}. Another approach with minimal interference to the MR acquisition is to use navigators based on highly-undersampled kSpace \cite{Ulrich2022}, which takes only $\SI{2.3}{ms}$ to detect the rigid-body movement.

Acquiring subject-specific additional data (e.g. as prescan) for the MRI reconstruction is common practice for conventional methods. E.g. the parallel imaging method GRAPPA requires ACS lines to reconstruct the missing kSpace points \cite{Griswold2002}. Similarly, SENSE requires measured coil sensitivity profiles to disentangle aliased MRI images \cite{Pruessmann1999}. Subject-specific NN were also proposed by Akçakaya et al. \cite{Ackaya2019} to perform MRI reconstructions.

Our investigation of the subject-specific NN fine-tuning suggested that the amount of the additional training datasets can be as little as three volumes. The amount of head positions used for training were investigated and no significant differences were shown in the range 3 to 6 datasets. However, there were no special instruction for the fine-tuning data. Further optimization could, thus, lead to improved results. The fine-tuning should be performed for a sufficient time, however after some point the improvement is saturated. In our case, 50 epochs (which took approximately 1 minute without any specific optimization) were sufficient for the fine-tuning training with the 6 head positions.

\subsection{Limitations and Outlook}

The paper presents a proof-of-principle and there are many details which can be improved. The training dataset was created from only 11 volunteers. For each volunteer only $B_0$ maps at 30 positions were acquired and although the distribution of test and training data was similar, more data on a much more diverse group of subjects (e.g., different head sizes) will be necessary to achieve optimal results. The training dataset was augmented by a physics-driven concept. However, the number of head positions remained unchanged. The augmentation, thus, simulates only the differences in the $B_0$ shimming of the volume-of-interest. Also other augmentation approaches and tests on generalizability should be considered. The subject-specific fine-tuning was performed with six volumes in 50 epochs. No special instruction were given, which can be improved by a tailored process of fine-tuning sampling for a given coil.   

Future research should involve a combination with external motion tracking hardware and evaluating the benefits for $B_0$- and motion-sensitive MRI sequences. Optical tracking could provide updates of the patient position with temporal resolution up to 85 Hz. The proposed method requires calibration sampling for the fine-tuning and short training. After that the input to the NN consist of anatomical images, the initial $B_0$ map from the beginning of the measurement protocol, and information of the rigid movement. The prediction of the $B_0$ map is completely independent from the MR scanner. Thus, the valuable information about the change of the $B_0$ map due to subjects movement, which is usually available only from lengthy volumetric navigators, could be predicted with the same temporal resolution as is available from rigid-body motion logs. In the future these predicted $B_0$ maps could even be used for real-time updating both the MRI volume and the $B_0$ shim parameters together during the acquisition of the MR data. This can ultimately lead to significantly improved data quality for a range of $B_0$-sensitive MRI methods.

\section{Conclusion}

This paper presents the proof-of-principle implementation of a new deep learning-based and subject-specific approach to predicting the change of the $B_0$ maps due to patient movement. Results were compared to the ground truth and the established EPI approach, which is equivalent to using lengthy volumetric navigators. Our results suggest that the prediction of $B_0$ maps is feasible and highly accurate. In combination with external tracking, a considerable improvement in data quality of $B_0$-sensitive MRI methods could be expected.

\bibliographystyle{ieeetr}
\bibliography{references}

\end{document}